\documentclass{article}

\usepackage{arxiv}

\usepackage[utf8]{inputenc} 
\usepackage[T1]{fontenc}    
\usepackage{hyperref}       
\usepackage{url}            
\usepackage{booktabs}       
\usepackage{amsfonts}       
\usepackage{amsmath}
\usepackage{amssymb}
\usepackage{dsfont}
\usepackage{nicefrac}       
\usepackage{microtype}      
\usepackage{cleveref}       
\usepackage{graphicx}
\usepackage[numbers]{natbib}
\usepackage{doi}
\usepackage{listings}

\usepackage{tabularray}
\usepackage{multirow}

\usepackage{pgfplots}
\DeclareUnicodeCharacter{2212}{−}
\usepgfplotslibrary{groupplots,dateplot}
\usetikzlibrary{patterns,shapes.arrows}
\pgfplotsset{compat=newest}

\def\etal{\emph{et al}}

\title{Transferability of Representations Learned using Supervised Contrastive Learning Trained on a Multi-Domain Dataset}

\date{}

\usepackage{authblk}

\author[1]{\hspace{1mm}Alvin De Jun Tan}
\author[1]{\hspace{1mm}Clement Tan}
\author[1]{\hspace{1mm}Chai Kiat Yeo}
\affil[1]{School of Computer Science and Engineering, Nanyang Technological University, Singapore}

\renewcommand{\headeright}{}
\renewcommand{\undertitle}{}
\renewcommand{\shorttitle}{}

\begin{document}
\maketitle

\begin{abstract}
Contrastive learning has shown to learn better quality representations than models trained using cross-entropy loss. They also transfer better to downstream datasets from different domains. However, little work has been done to explore the transferability of representations learned using contrastive learning when trained on a multi-domain dataset. In this paper, a study has been conducted using the Supervised Contrastive Learning framework to learn representations from the multi-domain DomainNet dataset and then evaluate the transferability of the representations learned on other downstream datasets. The fixed feature linear evaluation protocol will be used to evaluate the transferability on 7 downstream datasets that were chosen across different domains. The results obtained are compared to a baseline model that was trained using the widely used cross-entropy loss. Empirical results from the experiments showed that on average, the Supervised Contrastive Learning model performed 6.05\% better than the baseline model on the 7 downstream datasets. The findings suggest that Supervised Contrastive Learning models can potentially learn more robust representations that transfer better across domains than cross-entropy models when trained on a multi-domain dataset.
\end{abstract}


\section{Introduction}
\label{sec:intro}

In recent years, there has been renewed interest in contrastive learning due to its success in self-supervised learning for computer vision tasks. This has led to a resurgence of research related to contrastive learning and these research produced results comparable or even outperformed state-of-the-art results obtained for its supervised counterpart in the ImageNet benchmark \cite{chen_simple_2020, he_momentum_2020, chen_exploring_2021, khosla_supervised_2020, grill_bootstrap_2020, zbontar_barlow_2021}. Contrastive learning is a technique in representation learning that is used to learn an embedding space to represent the features from the dataset of interest. By contrasting samples, the aim is for sample pairs that are similar to stay close to each other and sample pairs that are dissimilar to stay far apart from each other in the embedding space. Motivated by the promising performance of self-supervised contrastive learning, Khosla \etal \cite{khosla_supervised_2020} proposed the Supervised Contrastive Learning framework,  which leveraged on the label information and incorporated the use of labels into their contrastive training objective, known as the SupCon loss. Supervised Contrastive Learning achieved a better performance in ImageNet accuracy using the proposed SupCon loss than the standard cross-entropy loss \cite{khosla_supervised_2020}.

Models trained using contrastive learning have been shown to provide comparable or even better quality representations than models trained using supervised cross-entropy loss, especially in the image classification task \cite{he_momentum_2020, chen_simple_2020, khosla_supervised_2020, tian_what_2020, zhao_what_2020, islam_broad_2021, zbontar_barlow_2021, grill_bootstrap_2020}. Islam \etal \cite{islam_broad_2021} showed that the representations learned from using contrastive objectives contained more low-level and/or mid-level semantics than cross-entropy models, allowing them to more effectively transfer the representations learned into a new task quickly. Previous studies have also shown that the representations learned can be easily transferable to a different downstream dataset \cite{islam_broad_2021, chen_simple_2020, khosla_supervised_2020, grill_bootstrap_2020, zbontar_barlow_2021}. However, the representations were often learned using a single source dataset that contains images from a single domain (e.g. ImageNet \cite{ILSVRC15}). There has been little work done that explored the transfer capability of learned representations from a multi-domain dataset using contrastive learning. Taking a step back from our discussion of contrastive learning, we note that Convolutional Neural Networks (CNN) trained on a single dataset from a single domain often end up learning biased and less robust representations, and will likely perform well only on the specific domain it was trained for \cite{torralba_unbiased_2011}.

\begin{figure}[t]
    \centering
    \includegraphics[width=0.9\textwidth]{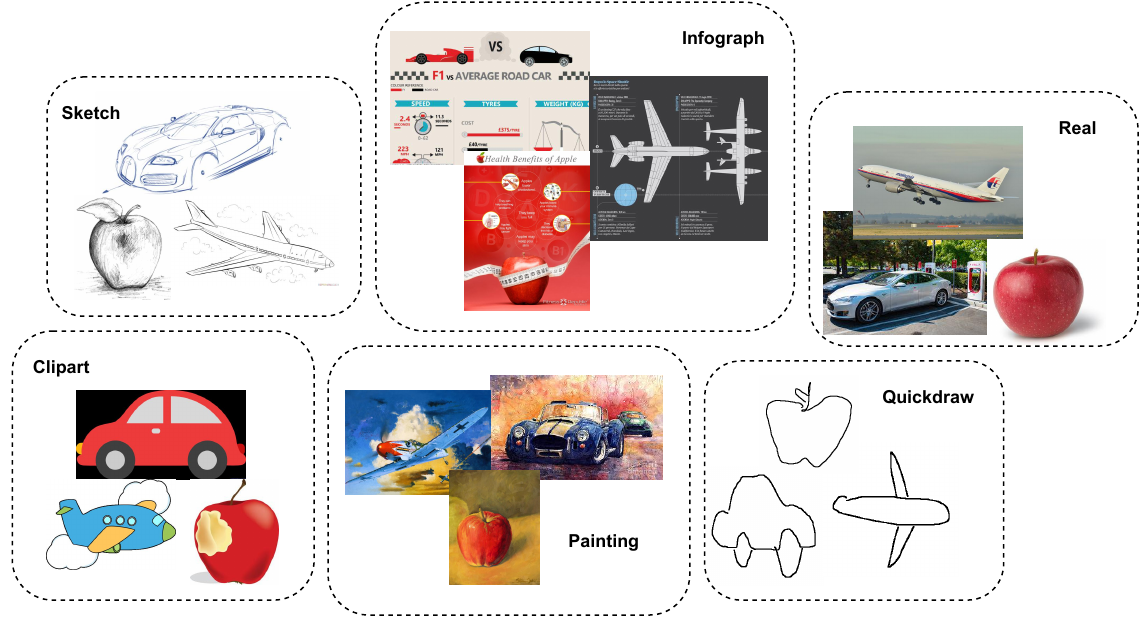}
    \caption{A Dataset Consisting of Images from Multiple Domains (Domains and Images were selected from DomainNet)}
    \label{fig:DomainNet dataset}
\end{figure}

Inspired by the superior capability of contrastive learning in providing better quality representations and to learn more robust representations that are able to generalize to downstream datasets across different domains, we applied Supervised Contrastive Learning on a multi-domain dataset. We compared the transfer performance of the learned representations to a model trained using supervised cross-entropy loss on 7 downstream datasets across different domains. The cross-entropy model forms the baseline in my study. To evaluate transfer performance, a fixed feature linear evaluation protocol was used. The multi-domain dataset used was DomainNet \cite{peng_moment_2019}. To the best of our knowledge, this is the first research to apply Supervised Contrastive Learning on DomainNet and evaluate the transferability of the learned representations on downstream datasets across different domains. More specifically, we seek to answer the following question in this study: does supervised contrastive learning give better transfer performance than the commonly used cross-entropy loss when trained on a multi-domain dataset?

Figure \ref{fig:DomainNet dataset} illustrates the DomainNet dataset consisting of images from multiple related domains. Each domain contains the same classes as the others. When training using the SupCon loss, the set of positives would contain images from the same classes but these images could be from different domains. These positives would be contrasted against the negatives, which are images of a different class. Using a multi-domain dataset can increase the data variety, potentially enriching the robustness of the representations learned and allowing knowledge transfer across domains. The main contributions of this paper can be summarized as follows:
\begin{itemize}
    \item We applied Supervised Contrastive Learning on a multi-domain dataset by combining all images from the different domains in DomainNet. 
    \item We compared the transfer performance of the representations learned from Supervised Contrastive Learning to the baseline: a model trained using the widely used cross-entropy loss.
    \item We evaluated the learned representations using the fixed feature linear evaluation protocol on 7 downstream datasets selected across different domains. Our results showed that the transfer performance of the Supervised Contrastive Learning model outperformed the baseline model on all the downstream datasets selected when trained on the combined DomainNet dataset. 
\end{itemize}

\section{Related Work}

{\bf Transfer Learning.} In general, for deep neural networks to perform well on a task, huge amount of data and compute power is required. As a result, using deep learning to train a different model for each task from scratch is expensive, in terms of the cost of obtaining a large amount of good quality task-specific data and the computing resources required. Transfer learning provides a solution to this, by training a model using a large-scale dataset (e.g. ImageNet), and then transferring the features learned to many downstream tasks. In this way, the downstream tasks can be trained with significantly lesser data. Earlier research showed that using features from ImageNet-trained models to train Support Vector Machines (SVM) and logistic regression classifiers outperformed manual hand-engineered features \cite{razavian_cnn_2014, donahue_decaf_2014, chatfield_return_2014}. Kornblith \etal \cite{kornblith_better_2019} showed that better ImageNet-trained models (in terms of accuracy) tend to provide better features for transfer learning and fine-tuning. Mensink \etal \cite{mensink_factors_2021} found that success of transfer depends on the source and target task types and the source dataset should include the domain of the target dataset for better transfer performance. Most of the previous research studied transfer learning in the premise of a model trained using a dataset comprising only a single domain (e.g. ImageNet). Moreover, the models studied in previous work were trained using cross entropy loss.

A similar study to our research was done by Islam \etal \cite{islam_broad_2021}. They studied the transferability of representations learned using contrastive learning on downstream datasets across different domains. They combined supervised and self-supervised contrastive training objectives with cross entropy loss to see if it can improve transfer performance. However, the models were also pre-trained using ImageNet.

Different from those previous works, in this paper, we studied the transfer performance of a model trained with Supervised Contrastive Learning and compared it to a model trained with cross-entropy loss using a multi-domain dataset. The transfer performance was evaluated using linear evaluation with fixed feature extractor on 7 downstream datasets that were selected from different domains.

\begin{figure}[t]
\centering
\begin{tabular}{cc}
\includegraphics[width=0.45\textwidth]{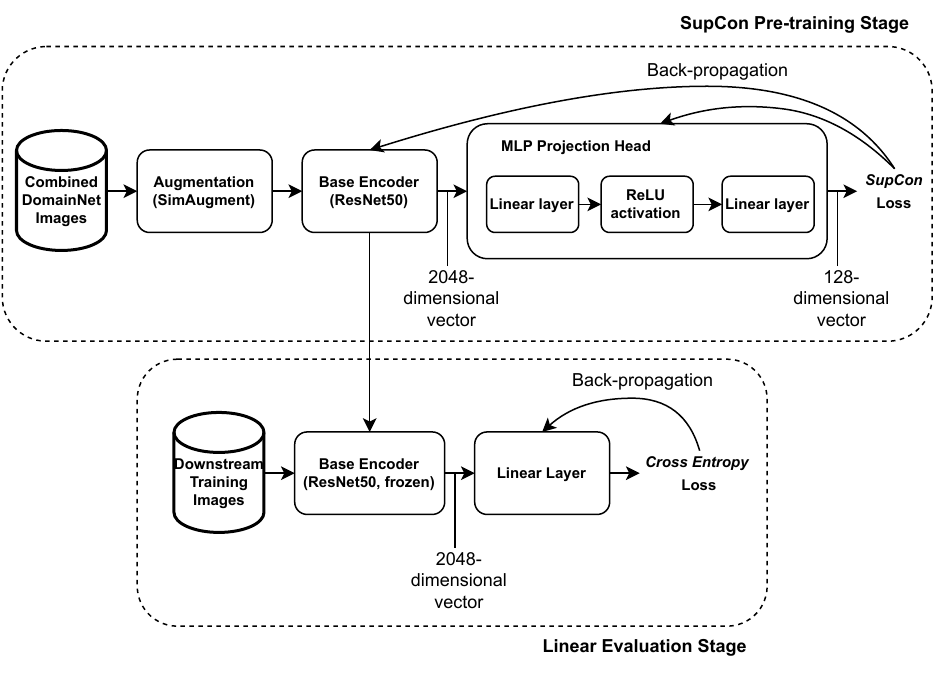}&
\includegraphics[width=0.4\textwidth]{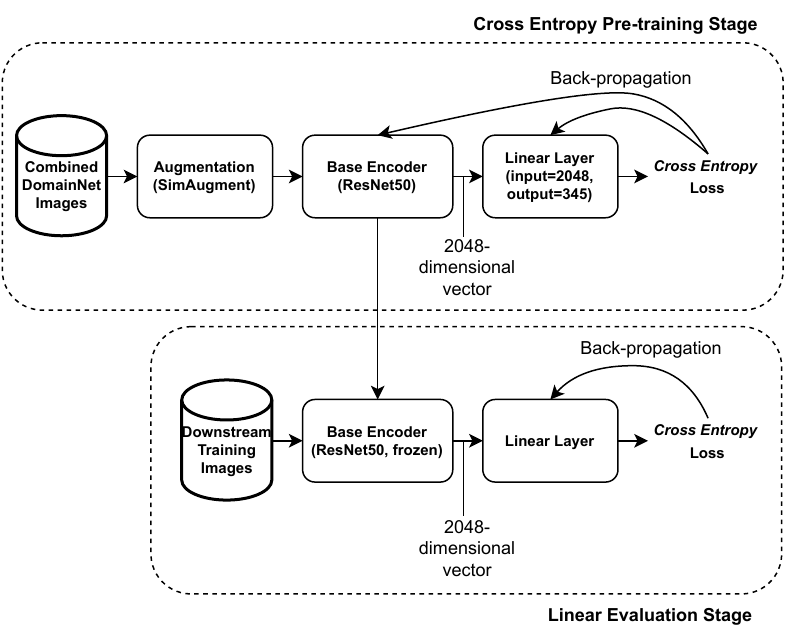}\\
(a)&(b)
\end{tabular}
\caption{Overview of Training Process for: (a) SupCon Model; (b) Cross-Entropy Model.}
\label{fig:Overview of Training Process}
\end{figure}

\section{Methodology}

\subsection{Analysis Setup}

Given $S$ number of source domains, $D_i = \{(\textbf{x}_{i1}, y_{i1}), (\textbf{x}_{i2}, y_{i2}), ..., (\textbf{x}_{iN}, y_{iN})\}$, where $i \in \{1, 2, ..., S\}$, each source domain has the same number of classes, and a marginal distribution of $P_i$. We have a target domain $D_T = \{(\textbf{x}_{T1}, y_{T1}), (\textbf{x}_{T2}, y_{T2}),$ 
$ ..., (\textbf{x}_{TN}, y_{TN})\}$, with marginal distribution of $P_T$. It is not necessary that $P_i = P_T$. The objective is to learn a target prediction function $f(\cdot)$ using all the knowledge from the source domains $D_1, D_2, ..., D_S$. We used the linear evaluation over fixed feature extractor as the target prediction function in this study. 

\subsection{Cross-Entropy Loss}
Given an input image $i$, $K$ number of classes, and \(s_{ij}\) being the logit for class \(j\), where \(j \in \{1, ..., K\}\), the multi-class cross-entropy loss is given by:
\begin{equation}
\mathcal{L}_i^{Cross-Entropy} = -\sum^{K}_{k=1}{t_i * log(\frac{e^{s_{ij}}}{\sum^{K}_{j'=1}{e^{s_{ij'}}}})}
\end{equation}
where \(t_i\) is the ground truth label for input \(i\), and evaluates to 1 if \(t_i = k\), 0 otherwise. This can be easily extended to a mini-batch setting, and the loss for the mini-batch is given by:
\begin{equation}
\mathcal{L}^{Cross-Entropy} = \sum^{N}_{i=1}{\mathcal{L}_i^{Cross-Entropy}}
\end{equation}
where \(N\) is the number of samples in the mini-batch.

\subsection{Supervised Contrastive Loss}

Given a mini-batch of $N$ samples, $\{\textbf{x}_l, y_l\}_{l=1}^{N}$,  and the corresponding mini-batch of $2N$ samples $\{\tilde{\textbf{x}}_l, \tilde{y}_l\}_{l=1}^{2N}$, where $\tilde{\textbf{x}}_{2k}$ and $\tilde{\textbf{x}}_{2k-1}$ are the two randomly augmented images of sample $\textbf{x}_{k}$, and $\tilde{y}_{2k} = \tilde{y}_{2k-1} = y_{k}$, where $k\in\{1, ..., N\}$. Denote the image encoder (e.g ResNet50) as $g(\cdot)$ and MLP projection head as $h(\cdot)$. $\textbf{z} = h(g(\textbf{x}))$ represents the projected vector representation of an image. The supervised contrastive loss (SupCon) is defined as follows:
\begin{equation}
\mathcal{L}^{SupCon}_i = \frac{-1}{2N_{\widetilde{\textbf{y}}_i} - 1}\sum_{j=1}^{2N} \mathds{1}_{i \neq j} \cdot 
\mathds{1}_{\widetilde{\textbf{y}}_i = \widetilde{\textbf{y}}_j} \cdot log\frac{exp(sim(\textbf{z}_i, \textbf{z}_j)/\tau)}{\sum_{k=1}^{2N}\mathds{1}_{i \neq k} \cdot exp(sim(\textbf{z}_i, \textbf{z}_k)/\tau)}
\end{equation}
\begin{equation}
\mathcal{L}^{SupCon} = \sum_{i=1}^{2N} \mathcal{L}^{SupCon}_i
\end{equation}

where,
\begin{itemize}
\item $sim(\cdot, \cdot)$ is the dot product between the two normalised vectors
\item subscript \(i\) denotes the index of an augmented image from the augmented mini-batch
\item  subscript \(j\) denotes the index of all the other augmented images from the augmented mini-batch that is not \(i\) and has the same class as \(i\) 
\item subscript \(k\) denotes the index of an augmented image from the augmented mini-batch that is not \(i\)
\item \(N_{\widetilde{\textbf{y}}_i}\) is the total number of images in the mini-batch (before augmentation) that has the same class as \(i\), hence \(2N_{\widetilde{\textbf{y}}_i} - 1\) is the number of augmented images in the augmented mini-batch that has the same class as \(i\) (note \(2N_{\widetilde{\textbf{y}}_i} - 1\) is also the number of images in the set of positives, consisting of all images with same class)
\end{itemize}

\begin{table}[t]
\centering
\caption{Summary of Datasets Used for Downstream Linear Evaluation}
\label{table:Summary of Datasets Used for Downstream Linear Evaluation}
\begin{tblr}{
  cell{2}{1} = {r=4}{},
  vlines,
  hline{1-2,6-9} = {-}{},
  hline{3-5} = {2-6}{},
}

Category & Dataset & {Train \\Size} & {Test \\Size} & {No. of \\Classes} & {Evaluation \\Metric} \\
Natural           & CIFAR10 \cite{krizhevsky_learning_2009}          & 50,000                           & 10,000                          & 10                                   & Top-1 Accuracy                          \\
                  & CIFAR100 \cite{krizhevsky_learning_2009}         & 50,000                           & 10,000                          & 100                                  & Top-1 Accuracy                          \\
                  & Flowers102 \cite{nilsback_automated_2008}      & 2,040                            & 6,149                           & 102                                  & {Mean-Per-Class \\Accuracy}             \\
                  & Aircraft \cite{maji_fine-grained_2013}        & 6,667                            & 3,333                           & 100                                  & {Mean-Per-Class \\Accuracy}             \\
Symbolic          & SVHN \cite{netzer_reading_2011}            &   73,257                               &   26,032                              &   10                                   &         Top-1 Accuracy                                \\
Illustrative      & Kaokore \cite{tian_kaokore_2020}         & 6542                             & 813                             & 8                                    & Top-1 Accuracy                          \\
Texutre           & DTD \cite{cimpoi_describing_2014}              & 3760                             & 1880                            & 47                                   & Top-1 Accuracy                          
\end{tblr}
\end{table}

\subsection{Datasets}

{\bf DomainNet.} DomainNet was used as the pre-training dataset. DomainNet [31] is a dataset consisting of common objects in six different domains: sketch, real, quickdraw, painting, infograph, clipart. Every domain includes 345 classes of objects, such as airplane, clock, flower and bus. For the purpose of this research, the images from all the different domains were combined into one huge dataset containing 409,832 images to form a multi-domain dataset. Note that only the training set split of DomainNet and the cleaned version was used\footnote{The dataset is available from http://ai.bu.edu/M3SDA/}. \newline 

\noindent {\bf Downstream datasets.} 7 downstream datasets were selected across different domains to evaluate the transferability of the representations learned. They were broadly categorized into natural, symbolic, illustrative and texture. The datasets were also selected with a mix of finer/coarser-grained labels, for e.g. CIFAR100, Aircraft and Flowers102 were finer-grained while CIFAR10 is coarser-grained. Table \ref{table:Summary of Datasets Used for Downstream Linear Evaluation} summarizes the downstream datasets. More details about the downstream datasets can be found in the supplementary materials. 

\subsection{Experimental Setup}

{\bf SupCon Model.} The SupCon training process was split into two stages (the pre-training stage and linear evaluation stage). Figure \ref{fig:Overview of Training Process}(a) shows an overview of the training process. In the pre-training stage, ResNet50 was used as the base encoder and the MLP projection head contains two linear layers (first linear layer was 2048-d with ReLU activation and the second linear layer was 128-d). Temperature $\tau = 0.13$  was used in the SupCon loss. The SGD optimizer with momentum of 0.9 and weight decay of 1e-4 was used to train the model, with a learning rate of 0.1. A batch size of 1024 was used and the model was trained for 400 epochs. The learning rate was warmed up linearly for the first 10 epochs, and we applied a step decay with decay rate of 0.1 at epochs 250 and 350 respectively. Further details on the augmentations used can be found in the supplementary materials.

In the linear evaluation stage, to evaluate the learned representations, the MLP projection head was discarded and a new linear layer was attached on top of the frozen ResNet50. The linear layer essentially acts as a linear classifier, with an input size of 2048 and output size corresponding to the number of classes that is in the downstream dataset to be trained on. The usual cross-entropy loss was used. No augmentation was applied on the training images other than resizing them to a dimension of 32x32 pixels. The SGD optimizer with momentum of 0.9 and no weight decay was used to optimize the cross-entropy loss. The linear layer was trained for a total of 50 epochs with no learning rate decay. We swept the hyperparameter space (learning rate and batch size) for each downstream dataset as follows:
\begin{itemize}
    \item Learning rate: 0.1, 0.01, 0.001
    \item Batch size: 32, 64, 128
\end{itemize}
The official training and test split was used for all the downstream datasets. If a training and validation split was available, they were used for the hyperparameter tuning, with the optimal learning rate and batch size selected from the model that gave the highest validation accuracy. If the dataset consists of multiple training and validation splits (e.g. DTD), we used the first split for hyperparameter tuning. If there was no training and validation split, we randomly selected 70\% of the training set for training and the remaining for validation. After hyperparameter tuning, the training and validation set was combined and used for training with the optimal learning rate and batch size, and the test accuracy was used to evaluate the transfer performance.
\newline 

\noindent {\bf Cross-Entropy (Baseline) Model.} The training process for the Cross-Entropy model is illustrated in Figure \ref{fig:Overview of Training Process}(b). No MLP projection head was used and the training objective to be optimized was the cross-entropy loss. A linear layer was attached on top of the base encoder, with input size of 2048 and output size of 345, which corresponds to the number of classes in the combined DomainNet dataset. The same hyperparameters as the SupCon model was used, except for a batch size of 512 and the learning rate was decayed at epochs 150, 250 and 350 respectively. In the linear evaluation stage, the same steps (including the hyperparameter tuning) as the SupCon model was performed.

\section{Results and Discussions}

\begin{table}[t]
\centering
\caption{Accuracy (\%) for SupCon and Cross Entropy Model on the Downstream Datasets for Linear Evaluation. (Note: Mean-per-class accuracy is provided for Aircraft and Flowers102 while the rest are top-1 accuracy. Mean and standard deviation over 5-runs are provided.)}
\label{table:accuracy for supcon and cross entropy model}
\resizebox{\textwidth}{!}{
\begin{tabular}{|l|c|c|c|c|c|c|c|c|}
\hline
Model                                                              & CIFAR10      & CIFAR100     & Aircraft              & Flowers102   & SVHN         & Kaokore      & DTD          & Mean  \\ \hline
SupCon                                                             & 92.31 ± 0.04 & 75.74 ± 0.05 & 36.53 ± 0.27          & 75.09 ± 0.09 & 70.03 ± 0.05 & 76.63 ± 0.2  & 51.26 ± 0.13 & \textbf{68.23} \\ \hline
\begin{tabular}[c]{@{}l@{}}Cross Entropy\\ (Baseline)\end{tabular} & 90.16 ± 0.06 & 70.99 ± 0.07 & 26.95 ± 0.35 & 65.92 ± 0.11 & 64.6 ± 0.07  & 71.22 ± 0.25 & 45.43 ± 0.3  & 62.18 \\ \hline
\end{tabular}
}
\end{table}

\begin{figure}[t]
\centering
    \resizebox{.5\linewidth}{!}{\input{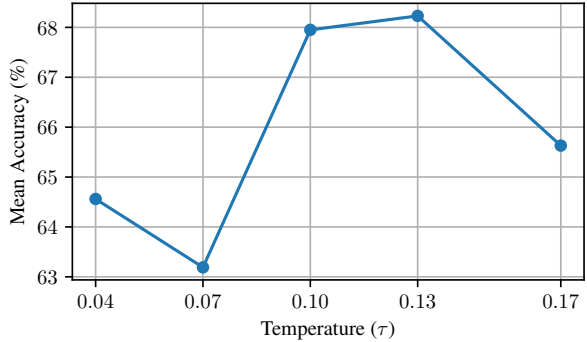}}
    \caption{Plot of Mean Accuracy over all the Downstream Datasets against Temperature}
    \label{fig:plot of mean accuracy against temperature}
\end{figure}

\begin{table}[t]
\caption{Accuracy (\%) for SupCon Models Trained with Different \(\tau\) Values. (Note: Mean-per-class accuracy is provided for Aircraft and Flowers102 while the rest are top-1 accuracy. Mean and standard deviation over 5-runs are provided. \(\tau = 0.13\) corresponds to the original model.)}
\label{table:ablation temperature}
\resizebox{\textwidth}{!}{
\begin{tabular}{|c|c|c|c|c|c|c|c|c|}
\hline
 Temperature ($\tau$) & CIFAR10      & CIFAR100     & Aircraft     & Flowers102   & SVHN         & Kaokore      & DTD          & Mean  \\ \hline
0.04                 & 90.66 ± 0.06 & 74.07 ± 0.1  & 28.51 ± 0.31 & 70.91 ± 0.09 & 65.41 ± 0.04 & 74.15 ± 0.18 & 48.25 ± 0.25 & 64.56 \\ \hline
0.07                 & 90.2 ± 0.01  & 72.48 ± 0.04 & 26.53 ± 0.31 & 67.06 ± 0.1  & 64.53 ± 0.04 & 74.07 ± 0.13 & 47.47 ± 0.05 & 63.19 \\ \hline
0.10                 & 91.9 ± 0.02  & 76.11 ± 0.04 & 35.27 ± 0.1  & 75.2 ± 0.09  & 70.01 ± 0.12 & 77.37 ± 0.13 & 49.83 ± 0.08 & 67.95 \\ \hline
0.13                 & 92.31 ± 0.04 & 75.74 ± 0.05 & 36.53 ± 0.27 & 75.09 ± 0.09 & 70.03 ± 0.05 & 76.63 ± 0.2  & 51.26 ± 0.13 & \textbf{68.23} \\ \hline
0.17                 & 91.85 ± 0.03 & 74.49 ± 0.07 & 31.98 ± 0.19 & 70.89 ± 0.12 & 67.55 ± 0.01 & 73.8 ± 0.08  & 48.81 ± 0.08 & 65.63 \\ \hline
\end{tabular}
}
\end{table}

\subsection{Linear Evaluation over Fixed Feature Extractor}

The linear evaluation was ran over 5 runs using the optimal set of learning rate and batch size found for each dataset and Table \ref{table:accuracy for supcon and cross entropy model} reports the average test accuracy along with its standard deviation. It can be observed that the SupCon model outperformed the cross entropy model in all the downstream datasets for linear evaluation with fixed feature extractor. The SupCon model obtained an average accuracy of 68.23\% among the downstream datasets, while the cross entropy model obtained an average accuracy of 62.18\%. The SupCon model performed, on average, 6.05\% better than the cross entropy model on the 7 downstream datasets when trained with a multi-domain dataset.

\subsection{Ablation Studies}

{\bf Effect of Temperature $\tau$.} The temperature ($\tau$) parameter used in the SupCon loss is adjustable and as noted in \cite{khosla_supervised_2020} that smaller $\tau$ values can benefit training more, but a very small value of $\tau$ can lead to numerical instability. A model trained with the optimal $\tau$ can improve its performance. As such, we studied the effect of $\tau$ on the transfer accuracy across the 7 downstream datasets for the SupCon model. Keeping other hyperparameters constant, we performed the experiments with different $\tau$ values. Figure \ref{fig:plot of mean accuracy against temperature} shows the plot of the mean accuracy against temperature. Table \ref{table:ablation temperature} shows the average accuracy and standard deviation (over 5 runs) obtained for each dataset. From Figure \ref{fig:plot of mean accuracy against temperature}, a similar trend as \cite{khosla_supervised_2020} was observed, where at lower temperature values of 0.04 and 0.07, the mean accuracy was lower than that of higher temperature values like 0.10 and 0.13. However, it is also observed that when temperature is increased further to 0.17, there is a drop in the mean accuracy. This would imply that it is important to select an optimal temperature value that can benefit the training process so that the representations learned can give better transfer performance. \newline

\noindent {\bf Effect of Augmentations.} As augmentations generally play an important role in contrastive learning \cite{chen_simple_2020, khosla_supervised_2020, zbontar_barlow_2021}, we studied the effect of augmentations on the transfer accuracy across the 7 downstream datasets for the SupCon model. In particular, we performed further experiments with AutoAugment (ImageNet policy), RandAugment and Stacked RandAugment. All the other hyperparameters were kept constant. Table \ref{table:ablation augmentations} shows the average accuracy and standard deviation (over 5 runs) obtained for each dataset. It can be observed that SimAugment and Stacked RandAugment, which are stronger augmentation strategies, performed better than AutoAugment (ImageNet policy) and RandAugment in terms of the mean accuracy obtained across all the downstream datasets, except for SVHN. We conjecture that in the SVHN dataset, the house numbers are often sheared or skewed, and the transformations in AutoAugment (ImageNet policy) and RandAugment include shearing, translation and rotation, which could potentially boost the transfer performance for SVHN. \newline

\noindent {\bf Effect of Base Encoder.} The base encoder acts as a feature extractor to extract useful representations of the underlying data. As such, the performance of the model is highly dependent on whether the base encoder can learn meaningful representations that can be transferred to downstream datasets. Hence, we decided to study the effect of using a deeper base encoder, ResNet101, on the transfer performance, with the premise that a deeper network has a larger capacity and can generalize better. Again, all the other hyperparameters were kept constant. Table \ref{ablation base encoder} shows the results obtained. However, we do not notice an improvement in the mean accuracy across the downstream datasets when ResNet101 was used. We conjecture that the combined DomainNet dataset that was used for pre-training was not large enough, hence even with a deeper network that has larger capacity, it was not able to learn better representations. Cross-referencing to a study done by Kolesnikov \textit{et al.} \cite{kolesnikov_big_2020} on the effect of model capacity and dataset size on transfer performance, it was found that the benefit from larger model diminishes given a constant number of pre-training images and solely increasing model capacity may degrade performance. Another plausible explanation could also be due to the small resolution (32x32) that the images were resized to during pre-training. A lower resolution may cause a lost in semantic information, of which a deeper network will not be able to take advantage of. Further studies on the effect of larger models on transfer performance may be required.

\begin{table}[t]
\caption{Accuracy (\%) for SupCon Models Trained with Different Augmentation Strategies (Note: Mean-per-class accuracy is provided for Aircraft and Flowers102 while
the rest are top-1 accuracy. Mean and standard deviation over 5-runs are
provided. SimAugment corresponds to the original model.)}
\label{table:ablation augmentations}
\resizebox{\textwidth}{!}{
\begin{tabular}{|l|c|c|c|c|c|c|c|c|}
\hline
Augmentation                                                            & CIFAR10      & CIFAR100     & Aircraft     & Flowers102   & SVHN         & Kaokore      & DTD          & Mean           \\ \hline
\begin{tabular}[c]{@{}l@{}}AutoAugment\\ (ImageNet Policy) \cite{cubuk_autoaugment_2019}\end{tabular} & 82.64 ± 0.02 & 59.5 ± 0.08  & 33.72 ± 0.15 & 52.77 ± 0.24 & 86.4 ± 0.04  & 70.58 ± 0.23 & 39.07 ± 0.19 & 60.67          \\ \hline
RandAugment \cite{cubuk_randaugment_2020}                                                            & 87.35 ± 0.03 & 67.52 ± 0.03 & 33.47 ± 0.19 & 66.67 ± 0.04 & 83.8 ± 0.1   & 74.42 ± 0.15 & 43.38 ± 0.16 & 65.23          \\ \hline
SimAugment \cite{chen_simple_2020}                                                              & 92.31 ± 0.04 & 75.74 ± 0.05 & 36.53 ± 0.27 & 75.09 ± 0.09 & 70.03 ± 0.05 & 76.63 ± 0.2  & 51.26 ± 0.13 & \textbf{68.23}          \\ \hline
\begin{tabular}[c]{@{}l@{}}Stacked\\RandAugment \cite{tian_what_2020}\end{tabular}           & 91.99 ± 0.02 & 76.04 ± 0.03 & 35.77 ± 0.21 & 75.35 ± 0.09 & 73.73 ± 0.09 & 75.94 ± 0.14 & 50.26 ± 0.05 & \textbf{68.44} \\ \hline
\end{tabular}
}
\end{table}

\begin{table}[t]
\centering
\caption{Accuracy (\%) for SupCon Models Trained with Different Base
Encoder (Note: Mean-per-class accuracy is provided for Aircraft and Flowers102
while the rest are top-1 accuracy. Mean and standard deviation over 5-runs are
provided. ResNet50 corresponds to the original model.)
}
\label{ablation base encoder}
\resizebox{\textwidth}{!}{
\begin{tabular}{|l|c|c|c|c|c|c|c|c|} 
\hline
Base Encoder & CIFAR10                           & CIFAR100                          & Aircraft                          & Flowers102                        & SVHN                             & Kaokore                           & DTD                               & Mean                        \\ 
\hline
ResNet50     & 92.31 ± 0.04                      & 75.74 ± 0.05                      & 36.53 ± 0.27                      & 75.09 ± 0.09                      & 70.03 ± 0.05                     & 76.63 ± 0.2                       & 51.26 ± 0.13                      & \textbf{68.23}              \\ 
\hline
ResNet101    & \multicolumn{1}{l|}{92.54 ± 0.02} & \multicolumn{1}{l|}{75.99 ± 0.05} & \multicolumn{1}{l|}{35.05 ± 0.14} & \multicolumn{1}{l|}{75.81 ± 0.01} & \multicolumn{1}{l|}{69.2 ± 0.07} & \multicolumn{1}{l|}{76.06 ± 0.26} & \multicolumn{1}{l|}{51.48 ± 0.26} & \multicolumn{1}{l|}{\textbf{68.02}}  \\
\hline
\end{tabular}
}
\end{table}

\subsection{Limitations}

Due to GPU memory constraints and that a large batch size was required for a good performance using the Supervised Contrastive Learning framework, the images in the combined DomainNet dataset were resized to a small resolution of 32x32 during pre-training. We acknowledge that this could affect the results as the lower image resolution would remove some semantic information that the model could potentially leverage on to learn better representations and boost transfer performance. It remains unknown if a larger image size such as 224x224 or 256x256 being used in pre-training with the DomainNet images may provide better transfer performance.

\section{Conclusion}

In conclusion, to answer the research question posed in the introduction, we empirically showed that supervised contrastive learning can give better transfer performance than cross entropy loss when trained on the multi-domain DomainNet dataset. Across the 7 downstream datasets that was selected, the SupCon model outperformed the cross entropy model in linear evaluation with fixed feature extractor by 6.05\% on average. The results from this study suggests that the representations learned from Supervised Contrastive Learning could perhaps be more robust and capture more domain invariant features that are transferable to downstream datasets across different domains. 

We would like to highlight an implication of the research on deep learning in the real world, where data distribution shift is a common problem faced. Data distribution shift refers to the phenomenon where the joint distribution of the inputs and outputs differ during the training and test stages. To give an example, an autonomous car that is trained on a dataset containing images of the icy roads in Norway would likely not perform well when put to test on the roads of Singapore. In safety critical systems such as autonomous driving, this could potentially result in catastrophic consequences. As such, it is important that we perform more research into building robust representations that can perform well across domains in the real world. Although the results of this research is far from that ultimate goal, we hope that this would inspire further research into the capabilities of contrastive training objectives or frameworks that are able to learn robust representations to be applied to downstream tasks across multiple domains.

\bibliographystyle{IEEEtranN}
\bibliography{references}  
\clearpage

\appendix
{\Large{\textbf{Supplementary Materials}}}

\section{Augmentations}
\label{sec:augmentations}

The following augmentation strategies were used in our experiments:
\begin{itemize}
    \item AutoAugment (ImageNet policy) \cite{cubuk_autoaugment_2019}: Using a reinforcement learning algorithm, the idea of AutoAugment was to search for an augmentation policy to optimize the performance for a dataset. In this case, the augmentation policy that was found to optimize the top-1 accuracy for ImageNet was used.
    \item RandAugment \cite{cubuk_randaugment_2020}: In contrast to AutoAugment, RandAugment eliminates the need for a search phase and uses random parameters to replace the parameters that were tuned by the reinforcement learning process in AutoAugment. 
    \item   SimAugment \cite{chen_simple_2020}: The same augmentation strategies used in SimCLR, which consist of random resized crop, random horizontal flip, color jitters and Gaussian blur.
    \item Stacked SimAugment \cite{tian_what_2020}: Combination of RandAugment and SimAugment. RandAugment was added as an additional step before color jitter in SimAugment.
\end{itemize}

The pseudocode (in PyTorch) are provided for reference in this section.

\subsection{AutoAugment (ImageNet policy)}

\begin{lstlisting}[breaklines=true]
from torchvision import transforms

# AutoAugment 
train_transform = transforms.Compose([
    transforms.Resize(size=32),
    transforms.AutoAugment(AutoAugmentPolicy.IMAGENET),
    # scale pixel values to between 0 and 1,
])
\end{lstlisting}

\subsection{RandAugment}
\begin{lstlisting}[breaklines=true]
from torchvision import transforms

# RandAugment
train_transform = transforms.Compose([
    transforms.Resize(size=32),
    transforms.RandAugment(),
    # scale pixel values to between 0 and 1,
])
\end{lstlisting}

\subsection{SimAugment}
\begin{lstlisting}[breaklines=true]
from torchvision import transforms

# SimAugment
train_transform = transforms.Compose([
    transforms.RandomResizedCrop(size=32, scale=(0.2, 1.)),
    transforms.RandomHorizontalFlip(),
    transforms.RandomApply([
        transforms.ColorJitter(0.4, 0.4, 0.4, 0.1)
    ]),
    transforms.RandomGrayscale(p=0.2),
    GaussianBlur(kernel_size=int(0.1 * 32)),
    # scale pixel values to between 0 and 1,
])
\end{lstlisting}

\subsection{Stacked RandAugment}
\begin{lstlisting}[breaklines=true]
from torchvision import transforms 

# Stacked RandAugment
train_transform = transforms.Compose([
    transforms.RandomResizedCrop(size=32, scale=(0.2, 1.)),
    transforms.RandomHorizontalFlip(),
    transforms.RandAugment(),
    transforms.RandomApply([
        transforms.ColorJitter(0.4, 0.4, 0.4, 0.1)
    ]),
    transforms.RandomGrayscale(p=0.2),
    GaussianBlur(kernel_size=int(0.1 * 32)),
    # scale pixel values to between 0 and 1,
])
\end{lstlisting}

\section{Downstream Datasets}

In this section, we briefly introduce the datasets that were used for downstream evaluation. 

\subsection{CIFAR10}
The CIFAR10 \cite{krizhevsky_learning_2009} dataset consists of 60000 images from 10 different classes. We used the CIFAR10 dataset as provided by PyTorch's torchvision module.

\subsection{CIFAR100}
The CIFAR100 \cite{krizhevsky_learning_2009} dataset has 60000 images from 100 different classes. We used the CIFAR10 dataset as provided by PyTorch's torchvision module.

\subsection{Aircraft}
Aircraft \cite{maji_fine-grained_2013} is a fine-grained dataset consisting of 10000 images from 100 different aircraft model variants. We used the FGVCAircraft dataset as provided by PyTorch's torchvision module.

\subsection{Flowers102}
Flowers102 \cite{nilsback_automated_2008} is a fine-grained dataset consisting of 8189 images of flowers from 102 different categories. We used the Flowers102 dataset as provided by PyTorch's torchvision module.

\subsection{SVHN}
SVHN \cite{netzer_reading_2011} consists of 99289 images of house numbers obtained from Google street view. There are 10 classes (corresponding to the digits from 0 to 9). We used the SVHN dataset as provided by PyTorch's torchvision module.

\subsection{Kaokore}
Kaokore \cite{tian_kaokore_2020} consists of 8848 face images from Japanese illustration (using version 1.1). However, there were some missing images, and after some data cleaning, we were left with 7355 images. There are two superclasses, ``gender" and ``status". The ``gender" class contains 2 subclasses, ``male" and ``female", while the ``status" class contains 4 subclasses, ``noble", ``warrior", ``incarnation", and ``commoner". Combining them, we obtained 8 classes. The images were downloaded from a Python script provided by the authors\footnote{The script is hosted on a GitHub repo: \url{https://github.com/rois-codh/kaokore}}.

\subsection{DTD}
DTD \cite{cimpoi_describing_2014} consists of 5640 texture images from 47 categories. We used the DTD dataset as provided by PyTorch's torchvision module.

\end{document}